\documentclass[runningheads]{llncs}

\usepackage[mobile]{eccv}

\newcommand{\ours}{WorldAgents}
\usepackage{eccvabbrv}

\usepackage{graphicx}
\usepackage{booktabs}

\usepackage[accsupp]{axessibility}  

\usepackage{inconsolata} 
\usepackage{listings}
\usepackage{xcolor}

\lstdefinestyle{mdStyle}{
    backgroundcolor=\color{gray!5},   
    basicstyle=\ttfamily\scriptsize,  
    breakatwhitespace=false,         
    breaklines=true,                  
    captionpos=b,                     
    keepspaces=true,                 
    numbers=none,                     
    showspaces=false,                
    showstringspaces=false,
    showtabs=false,                  
    frame=single,                     
    rulecolor=\color{gray!40},         
    breakindent=0pt,                  
    breakautoindent=false,            
}

\usepackage{hyperref}

\usepackage{orcidlink}

\begin{document}

\title{\ours{}: Can Foundation Image Models be Agents for 3D World Models?\vspace{-10px}} 

\titlerunning{\ours{}}

\author{Ziya Erko\c{c} \and
Angela Dai \and
Matthias Nie{\ss}ner \vspace{-5px}}

\authorrunning{Z.~Erko\c{c} et al.}

\institute{Technical University of Munich \\ \vspace{5px}
\url{https://ziyaerkoc.com/worldagents} \vspace{-5px}}

\maketitle
\begin{figure}
    \centering
    \scalebox{1.0}{%
        \begin{minipage}{0.88\textwidth}
    \includegraphics[width=\linewidth]{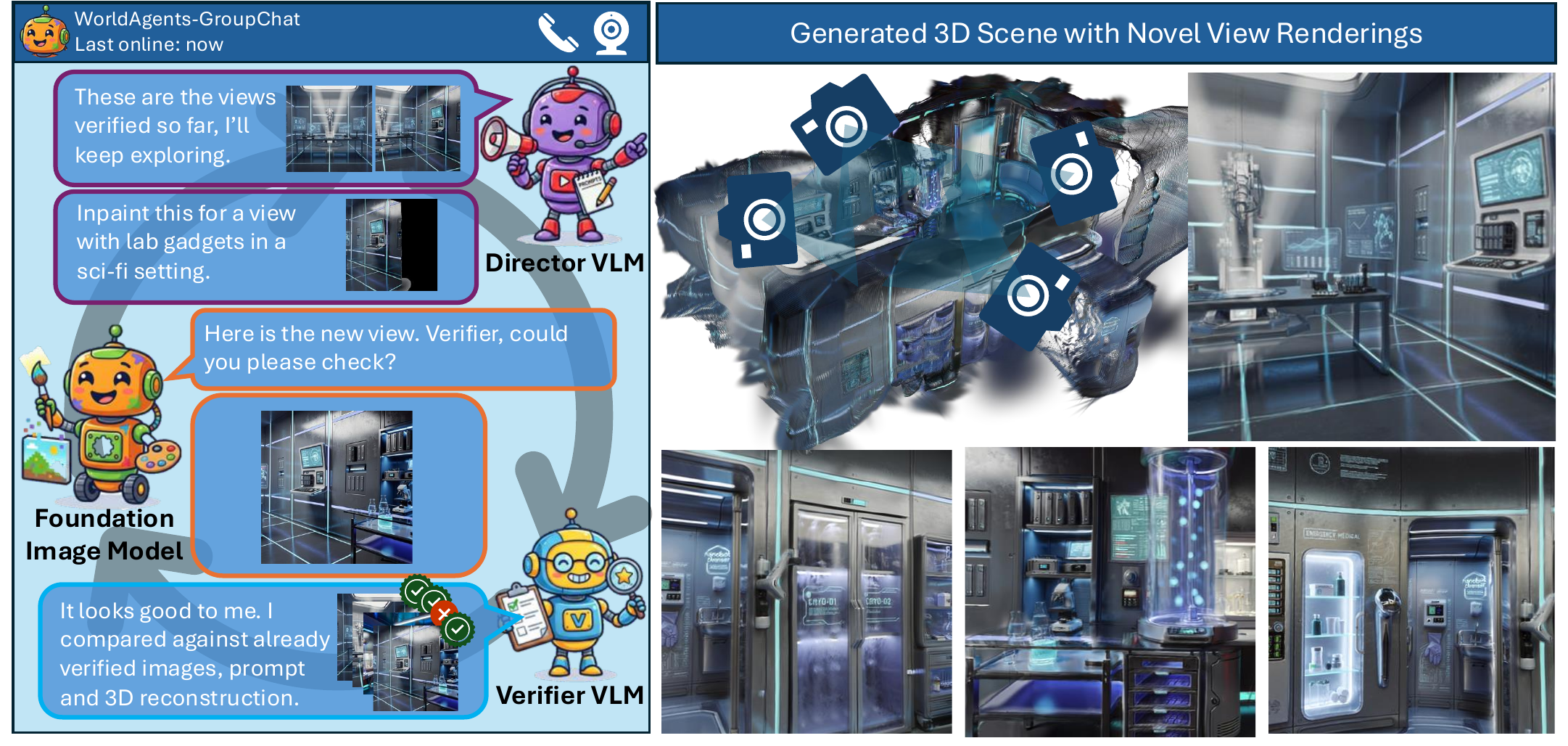}
    \caption{\textbf{\ours} employs 2D foundation models as agents in an iterative process to extract realistic, coherent 3D scenes from their learned distributions. 
    The scene generation process is orchestrated by a VLM agent (director) that generates prompts for new views to synthesize, fulfilled by an image generation model that creates these views, and verified by a VLM agent that assesses both 2D and 3D consistency of the new synthesized image(s).
    The output is a sci-fi lab. scene reconstructed with 3DGS, and can be visualized from novel views for exploration.}
    \vspace{-15px}
    \label{fig:teaser}
    \end{minipage}
    }
\end{figure}
\begin{abstract}
Given the remarkable ability of 2D foundation image models to generate high-fidelity outputs, we investigate a fundamental question: \textit{do 2D foundation image models inherently possess 3D world model capabilities?} 
To answer this, we systematically evaluate multiple state-of-the-art image generation models and Vision-Language Models (VLMs) on the task of 3D world synthesis. 
To harness and benchmark their potential implicit 3D capability, we propose an agentic framing to  facilitate 3D world generation. 
Our approach employs a multi-agent architecture: a VLM-based director that formulates prompts to guide image synthesis, a generator that synthesizes new image views, and a VLM-backed two-step verifier that evaluates and selectively curates generated frames from both 2D image and 3D reconstruction space. 
Crucially, we demonstrate that our agentic approach  provides coherent and robust 3D reconstruction, producing output scenes that can be explored by rendering novel views. Through extensive experiments across various foundation models, we demonstrate that 2D models do indeed encapsulate a grasp of 3D worlds. By exploiting this understanding, our method successfully synthesizes expansive, realistic, and 3D-consistent worlds.
\end{abstract}
\section{Introduction}
\label{sec:intro}
Recent rapid advances in 2D foundation models have revolutionized the field of computer vision. Text-to-image diffusion models demonstrate an unprecedented ability to generate high-fidelity, photorealistic images and exhibit deep semantic understanding of visual scenes~\cite{flux-2-2025, rombach2022high, esser2024scaling, baldridge2024imagen}. Trained on internet-scale datasets, these models encapsulate vast amounts of visual knowledge. While 2D generation has reached remarkable heights, the synthesis of immersive, 3D-consistent environments, often referred to as 3D world generation, remains a formidable challenge. Existing 3D generation methods~\cite{xiang2025structured3dlatentsscalable, tang2024diffuscene,siddiqui2023meshgptgeneratingtrianglemeshes, feng2023layoutgpt, chen2024meshanythingartistcreatedmeshgeneration, meng2025lt3sd, bokhovkin2024scenefactorfactoredlatent3d} are frequently bottlenecked by the scarcity of diverse, high-quality 3D training data or the computational complexity of maintaining multi-view consistency through Score Distillation Sampling~\cite{poole2022dreamfusiontextto3dusing2d, lin2023magic3dhighresolutiontextto3dcontent, wang2023prolificdreamerhighfidelitydiversetextto3d, tang2024dreamgaussiangenerativegaussiansplatting}.

Since 2D foundation models are trained on billions of 2D images, each of which represent 2D projections of our 3D spatial world, a compelling hypothesis emerges: these models may have implicitly learned the underlying spatial structures and physical rules of the environments they depict. This leads us to investigate a fundamental question: \textit{do 2D foundation image models inherently possess 3D world model capabilities?} If these models in fact learn a robust prior of the 3D world, they could theoretically be leveraged to bypass the reliance on explicit 3D datasets, serving as powerful engines for 3D scene synthesis. 

To answer this question, we systematically evaluate the implicit 3D spatial understanding of various state-of-the-art image generation models and VLMs. 
However, high-fidelity 3D reconstruction demands near pixel-perfect cross-view consistency, which single-pass prompting of a 2D model typically fails to guarantee.
To harness and benchmark the potential implicit 3D capabilities of such 2D  models, we propose a novel agentic method designed to orchestrate 2D foundation models for the task of consistent 3D world generation.

We thus cast 3D scene generation as a multi-agent process, comprising three specialized agents that work together to harness 2D image generation models to reconstruct coherent 3D worlds:
\begin{enumerate} 
\item \textit{VLM Director}, which acts as the high-level planner, dynamically formulating prompts to guide each new image generation and dictating the semantic evolution of the scene.
\item \textit{Image Generator}, which employs a 2D image generation model that executes spatial navigation by sequentially inpainting to synthesize novel, geometrically aligned views. Since the image models do not have explicit control of the camera positions, we applied inpainting approach to guide image model to complete the scene by dictating what to paint. 
\item \textit{VLM 2-Stage Verifier}, which serves as the critical quality control mechanism. Unlike standard rigid pipelines, this verifier provides fine-grained evaluation to selectively keep or discard generated frames. Crucially, it assesses consistency in two distinct stages: first from the 2D image space only, for semantic and structural coherence, and then  from the 3D reconstruction space to guarantee strict geometric alignment.
\end{enumerate}

We found that our agentic approach yields robust 3D reconstructions, allowing us to freely explore various generated environments by rendering arbitrary novel views. Through extensive experiments, we demonstrate that 2D foundation models do, in fact, encapsulate a profound grasp of 3D worlds. By exploiting this latent understanding through our carefully designed multi-agent orchestration, our method overcomes the limitations of independent 2D generation to successfully synthesize expansive, realistic, and strictly 3D-consistent worlds. In summary, our main contributions are as follows:
\begin{itemize}
    \item We provide a comprehensive investigation into the implicit 3D world model capabilities of state-of-the-art 2D image generation models guided by VLMs.
    \item We introduce a multi-agent architecture comprising a VLM director, a view generator, and a two-step verifier, specifically designed to harness 2D models for consistent 3D synthesis.
\end{itemize}

\section{Related Works}
\label{sec:related}

\subsection{3D World and Scene Generation}
There has been recent focus in building 3D worlds from text prompt or input views \cite{schneider2025worldexplorer, hollein2023text2room, 
zhang2026worldstereobridgingcameraguidedvideo, zhou2025stable, garcin2026pixelhistoriesworldmodels,chen2025flexworldprogressivelyexpanding3d,yang2025layerpano3dlayered3dpanorama, bahmani2026lyra}.
In particular, various methods have been proposed to leverage the powerful generative capacity of image models or video diffusion models, coupled with 3D-based control, typically through camera controlled conditioning. A line of work approaches the problem in the form of a panorama image generation~\cite{yang2025layerpano3dlayered3dpanorama,zhou2024dreamscene360unconstrainedtextto3dscene}. LayerPano3D~\cite{yang2025layerpano3dlayered3dpanorama} employs a layered image generation task. They fine-tune Flux~\cite{flux2024} to generate panorama images from text input. Unseen regions of each layers are inpainted with the same fine-tuned model. Additionally, DreamScene360~\cite{zhou2024dreamscene360unconstrainedtextto3dscene} includes a text-to-image diffusion model to generate a panorama image with a self-refinement mechanism using a VLM. To fully leverage existing image diffusion models, our approach operates entirely without additional pre-training. Furthermore, we introduce a multi-agent framework that actively guides the entire generation pipeline, rather than acting solely as a verifier. Crucially, our verification process ensures consistency directly within the 3D reconstruction space, moving beyond standard 2D image-domain validation. Our generation process includes iterative inpainting that does not require generating panorama images.

One line of work approaching this problem with both image- and depth-inpainting~\cite{yu2025wonderworld,hollein2023text2room}. Both WonderWorld~\cite{yu2025wonderworld} and Text2Room~\cite{hollein2023text2room} are using hand-crafted prompts to synthesize new regions to create 3D scenes. In contrast, we do not use handcrafted prompts but rely on VLM-based agents to orchestrate the scene generation process to construct navigable 3D scenes. Additionally, we employ an iterative, image-based inpainting strategy for scene generation to demonstrate the high degree of 3D consistency achievable without relying on explicit depth inpainting. Another line of work in scene generation includes retrieval-based layout-generation methods~\cite{feng2023layoutgpt, sun2025layoutvlmdifferentiableoptimization3d, tang2024diffuscene,lin2024instructsceneinstructiondriven3dindoor,yang2024physcenephysicallyinteractable3d, yang2024holodecklanguageguidedgeneration}. These methods rely on 3D layout data to be trained on which is orders-of-magnitude less than what image models have.
Major direction in video-based scene generation is camera-controlled models~\cite{bahmani2025ac3danalyzingimproving3d, bahmani2025vd3dtaminglargevideo, zhou2025stable}. Following advances in video synthesis, Stable Virtual Camera~\cite{zhou2025stable} demonstrated scene navigation and traversal through fine-tuning video diffusion models to provide camera-controlled multi-view generation. This creates compelling novel-view synthesis that can be used for further 3D reconstruction. WorldExplorer~\cite{schneider2025worldexplorer}, took this approach further to generate large 3D scenes that can be 3D reconstructed and arbitrarily rendered from novel views. Unlike these approaches, we employ VLM-based agents to guide the process and verify the generated frames without requiring crafting of a trajectory generation process. Our approach does not use any fine-tuned camera-controlled model but relies on existing text- and image-to-image 2D foundation models.  

\subsection{2D Foundation Image Models}
Past years have seen unprecedented advances in 2D image generation models~\cite{hu2024snapgen,flux-2-2025,team2023gemini,rombach2022high,peebles2023scalable,esser2024scaling}. Recent models can be conditioned on both text and multiple-images at the same time to achieve text-conditioned editing capabilities. Various methods have thus been built on top of these models to explore the capabilities for other downstream tasks such as 3D reconstruction using SDS (Score Distillation Sampling) and personalization~\cite{ruiz2023dreambooth,gal2022image, raj2023dreambooth3d,ruiz2024hyperdreamboothhypernetworksfastpersonalization, poole2022dreamfusiontextto3dusing2d}. Such image foundation models show strong capabilities in those downstream tasks. Their powerful generative and perception capacity have also inspired our approach to leverage  image foundation models to their full extent to understand if they would generate 3D-consistent views.   NanoBanana~\cite{team2023gemini} and Flux.2~\cite{flux-2-2025} some of the most recent methods that can generate high-fidelity images within a few seconds. We aim to exploit the full power of those image synthesis models to generate traversable 3D scenes. 

\subsection{Agent-Driven Generation and VLM Evaluators}
Recently, agent-based methods have achieved remarkable success across various domains~\cite{yin2026vision, jain2026nerfifymultiagentframeworkturning, feng2023layoutgpt, sun2025layoutvlmdifferentiableoptimization3d, deng2026humanobjectinteractionautomaticallydesigned}. These approaches leverage the robust visual and textual reasoning capabilities of Vision-Language Model (VLM) agents to tackle diverse tasks. Closest to our work is VIGA~\cite{yin2026vision}, which translates images into 3D scenes by generating corresponding Blender~\cite{blender} code. Their experiments demonstrate that VLMs possess a deep semantic understanding of scenes and can effectively manipulate code representations for image-to-3D reconstruction. Inspired by the strong reasoning capabilities of VLMs and recent advancements in 2D foundation models, we introduce a method for frame-by-frame 3D world generation. Unlike VIGA, which relies on a proxy code representation for static 3D reconstruction, our method directly generates image frames, and our ultimate objective is the synthesis of interactive, navigable 3D worlds from text prompts. 
\section{Method}
\begin{figure}
    \centering
    \includegraphics[width=\linewidth]{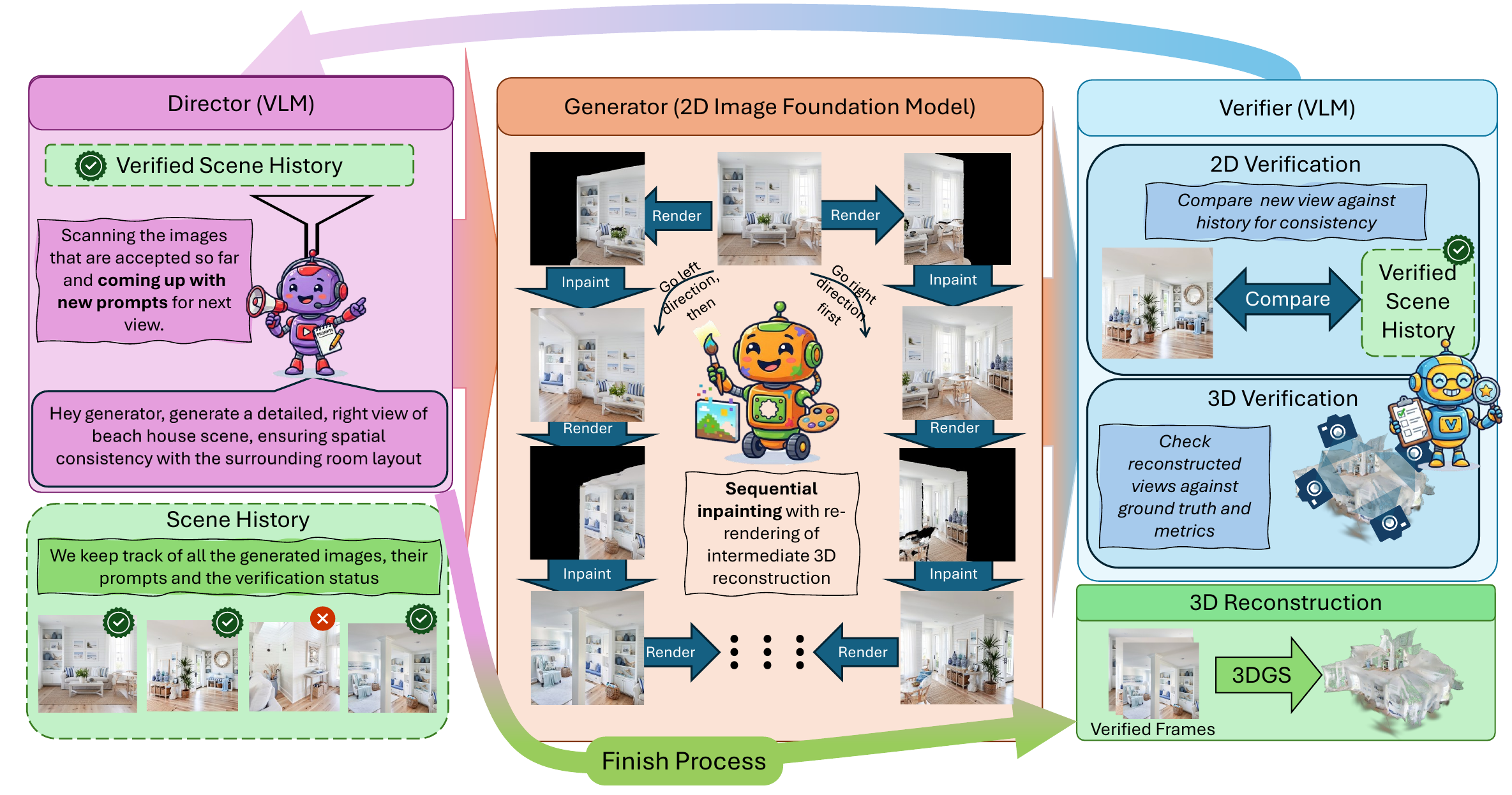}
    \caption{\textbf{Method overview.} Our method employs a multi-agent approach comprising a Director, a Generator, and a Verifier to construct coherent 3D scenes using image diffusion models. The Director guides the overall process by formulating novel prompts. The Generator then leverages sequential inpainting to synthesize 3D-consistent views. Subsequently, the Verifier evaluates these generated views to ensure rigorous multi-view consistency. Finally, the verified frames are reconstructed into a 3D Gaussian Splatting (3DGS) representation.}
    \label{fig:overview}
\end{figure}

Figure~\ref{fig:overview} presents an overview of our proposed method. We formulate 3D scene generation as a collaborative process orchestrated by three specialized agents: a Generator, a Verifier, and a Director. The Generator is a 2D image foundation model capable of text- and image-conditioned synthesis; we leverage it to inpaint specific regions based on scene captions generated by Director. To maintain global consistency, the Verifier evaluates each newly generated image against a history of previously accepted views. It concurrently maintains an intermediate 3D reconstruction to ensure the generated views form a geometrically coherent 3D space. The overall iterative process is guided by the Director, which analyzes the verified view history to propose descriptive prompts for novel viewpoints. Once the Director determines that the scene is comprehensively covered, the generation process terminates, and the accumulated views are utilized to reconstruct the final 3D Gaussian Splatting (3DGS) representation using AnySplat~\cite{jiang2025anysplat}.
\label{sec:method}

\subsection{Problem Formulation}

Given an input text description $y_1$, our aim is to generate a spatially coherent 3D scene $\mathcal{W}$ representative of $y_1$.
Concretely, we produce a set of $N$ posed images that collectively define a 3D world and can be reconstructed as 3D Gaussians \cite{kerbl3Dgaussians} to enable navigation and exploration of the 3D world. We provide agents with total of $\hat{R}$ number of tries to generate $N$ images. 
We additionally include the text description for each frame in our world state. In total, our world state is composed of a series of verified 2D image, their camera poses and corresponding text prompt, which are acquired through agentic process employing 2D foundation models and VLMs.
Formally, we represent this scene as:
$$
\mathcal{W} = \{(I_i, P_i, y_i)\}_{i=1}^{N}
$$
where each $I_i$ is a high-fidelity image and $P_i$ is its corresponding absolute camera pose in the global coordinate system. The initial frame generation step is a special case, as it does not involve Director agent. It is just a text-to-image generation task using $y_1$.

We formulate 3D world generation as an iterative, agent-directed process. At each discrete time step $t$, a director agent $\mathcal{D}$ analyzes the current world state $\mathcal{W}_t = \{(I_i, P_i, y_i)\}_{i=1}^{t}$ to propose how to expand the region by generating text prompt $y_t$. When the overall generation has gone through $\frac{\hat{R}}{2}$ tries, it switches to exploring left. Next camera view is calculated as $P_{t+1} = P_t \circ \Delta P_t$, where $\Delta P_t$ is a relative transformation of camera towards either right- or left-direction in a fixed amount, it also contains a random perturbation to create more diverse coverage for the next view.

Given the previous view $I_t$ and the new camera pose $P_{t+1}$, a  generator agent $\mathcal{G}$ relies on a 2D foundation model to synthesize a candidate view $\hat{I}_{t+1}$ 

As 2D foundation models can be prone to structural hallucinations that violate multi-view geometry constraints, we introduce a strict 2-Stage Verifier $\mathcal{V}$. The Verifier acts as a binary gating function that evaluates the candidate $\hat{I}_{t+1}$ against the established world $\mathcal{W}_t$ across both 2D semantic space and 3D reconstruction space:
$$
\mathcal{V}(\hat{I}_{t+1}, \mathcal{W}_t) = 
\begin{cases} 
1, & \text{if } \hat{I}_{t+1} \text{ is semantically and geometrically consistent with $\mathcal{W}_t$} \\ 
0, & \text{otherwise} 
\end{cases}
$$

The candidate view is appended to the global state (i.e., $\mathcal{W}_{t+1} = \mathcal{W}_t \cup \{(\hat{I}_{t+1}, P_{t+1}, y_{t+1})$) if and only if $\mathcal{V}(\hat{I}_{t+1}, \mathcal{W}_t) = 1$. If rejected, the candidate is discarded, and the generation step is re-sampled. By optimizing this discrete acceptance criteria, our approach guarantees that the final generated world $\mathcal{W}$ adheres to multi-view constraints while exploiting the superior visual fidelity of the underlying 2D foundation model.

The process ends when we hit $\hat{R}$ maximum number of images, or the director agent concludes that all of the scene is observed and gives stop signal.
\subsection{Director Agent}
The Director agent, denoted as $\mathcal{D}$, serves as the semantic orchestrator of the 3D world synthesis process. To prevent the semantic drift and unconstrained wandering typical of autoregressive video generation, $\mathcal{D}$ dynamically computes the next logical viewpoint based on the exploration history.

At each time step $t$, the Director observes the current state of the generated world $\mathcal{W}_t$ alongside the overarching global text prompt $y$. It is parameterized by a Vision-Language Model (VLM) that acts as a policy, mapping this environmental context to a view-specific text prompt $y_t$ by checking out the world state and the previous prompts: $y_{t+1} = \mathcal{D}(\mathcal{W}_t)$.

Concurrently, $y_{t+1}$ explicitly defines the expected visual content from the new perspective, providing strict semantic conditioning for the Generator. It includes textual description of where to investigate and what to be included in that part of the scene when the camera pose changes. The $y_1$ is already provided as an input, therefore, the Director agent is not involved in the generation of first frame.

By prompting the VLM to iteratively predict $y_{t+1}$, our framework functions as an autonomous, context-aware semantic operator, ensuring that the exploration trajectory creates meaningful scenes strictly aligned with the global semantic prior $y$. For instance, our director agent suggest following prompts in sci-fi scene for one iteration \textit{``expand further right, seamlessly continuing the sleek metallic wall panels ... wrapping blue and cyan neon strips ... a large, translucent cylindrical containment unit with softly pulsing blue lights ... embed a recessed digital control panel"}. It provides comprehensive and semantically-rich prompts for the next view by also keeping the overall context in the sci-fi scene.

Our trajectory procedure starts from the first frame and first goes in the right direction and then the left direction. We prompt Director about which direction we are heading now. After that, we apply fixed rotation of $\phi$ degrees around the up-axis to form $R_{\text{fixed}}$. To increase coverage diversity we apply random transformation, $T_{\text{random}}$ on top of that.
$$\underbrace{P_{t+1}}_{4 \times 4} = \underbrace{T_{\text{random}}}_{4 \times 4} \cdot \underbrace{R_{\text{fixed}}}_{4 \times 4} \cdot \underbrace{P_t}_{4 \times 4}$$

We calculate $P_{t+1}$ using that formula, where $\cdot$ means matrix multiplication, and $\Delta P_t = T_{\text{random}} \cdot R_{\text{fixed}}$. If the process has gone through $\frac{\hat{R}}{2}$ tries, the director switches the process to exploring left of the initial frame.

\subsection{Generator Agent}

The Generator agent, $\mathcal{G}$, is tasked with synthesizing a high-fidelity candidate view $\hat{I}_{t+1}$ that adheres to the  the semantic conditioning $y_{t+1}$ provided by the Director and geometric transformation $P_{t+1}$. 
To embed 3D structure and camera awareness into the 2D generation process, we reinterpret the 2D generative model as  sequential inpainting. 
Each new image is conditioned on re-rendered views based on the reconstruction from previously generated views, ensuring geometric consistency across the scene.

In order to generate a new image, we first collect the $\mathcal{W}_t$ to reconstruct a 3DGS scene. We then re-render the scene from a new view to provide it as an input to our image diffusion model. Specifically, we utilize AnySplat~\cite{jiang2025anysplat} to lift $\mathcal{W}_t$ into a global set of 3D Gaussians, denoted as $\Theta_t$:
$$
\Theta_t = \mathcal{F}_{\text{AnySplat}}(\mathcal{W}_t)
$$

To continue exploring the environment and synthesize the subsequent novel view, we compute a target camera pose $P_{t+1}$, which includes fixed rotation towards either left- or right-direction. Rather than relying on a strictly deterministic trajectory, we introduce a stochastic exploration mechanism by applying a randomly perturbed transformation so that the generator can get a more diverse coverage of the scene. 
Finally, we employ the Gaussian rasterizer $\mathcal{R}$ to render the reconstructed scene from the  novel viewpoint $P_{t+1}$. This yields the rendered image $I^{\text{warp}}_{t+1}$:
$$
I^{\text{warp}}_{t+1} = \mathcal{R}(\Theta_t, P_{t+1})
$$

Due to camera translation and rotation, $I^{\text{warp}}_{t+1}$ inevitably contains missing regions caused by disocclusions and new camera field of view. We leverage a pre-trained 2D foundation model, $\mathcal{G}_{\text{inpaint}}$, to complete the missing visual information. The model is conditioned on the known warped pixels, and the localized text prompt $y_t$ from director:
$$
\hat{I}_{t+1} = \mathcal{G}_{\text{inpaint}}(I^{\text{warp}}_{t+1}, y_{t+1})
$$

By grounding the generation process in explicit 3D reprojection before applying the 2D generative prior, the Generator ensures that the overlapping regions between $I_t$ and $\hat{I}_{t+1}$ remain geometrically rigidly aligned, while the foundation model is constrained purely to filling in the structurally logical, disoccluded regions.

\subsection{2D \& 3D Verifier Agents}
Because 2D foundation models are prone to structural hallucinations and perspective distortions, we introduce a rigorous 2-Stage Verifier agent, $\mathcal{V}$, to act as a definitive gating mechanism for which images should compose the 3D world.
The Verifier ensures that the candidate view $\hat{I}_{t+1}$ is both semantically aligned with the Director's intent and strictly geometrically consistent with the established 3D world $\mathcal{V}_{t}$. The verification is decomposed into a 2D semantic check and a 3D reconstruction-space check.

\paragraph{Image-Space Verification}  
First, we employ a Vision-Language Model (VLM) to assess the semantic coherence and visual quality of the candidate image. The VLM, denoted as $\mathcal{V}_{\text{2D}}$, takes the candidate view $\hat{I}_{t+1}$, the world state $\mathcal{W}_t$, and the director's prompt $y_t$ to detect obvious visual artifacts, domain shifts, or prompt misalignment. The output is a binary decision: $
v_{\text{2D}} = \mathcal{V}_{\text{2D}}(\hat{I}_{t+1}, \mathcal{W}_t) \in \{0, 1\}$.

\paragraph{3D Reconstruction-Space Verification}
Even if a candidate frame is semantically plausible in 2D, it may harbor subtle geometric distortions that violate multi-view consistency. To enforce strict global 3D consistency, we assess how the introduction of the candidate view $\hat{I}_{t+1}$ impacts the overall integrity of the 3D reconstruction of the scene. We define a provisional global state $\mathcal{W}'_{t+1} = \mathcal{W}_t \cup (\{\hat{I}_{t+1}, \hat{P}_{t+1}\})$, representing all verified frames up to step $t$ plus the new candidate. We lift this provisional set into a unified 3D representation using AnySplat~\cite{jiang2025anysplat}, yielding a candidate 3DGS model: $\Theta'_{t+1} = \mathcal{F}_{\text{AnySplat}}(\mathcal{V}'_{t+1})$. 

To quantify the global structural integrity, we render the scene from all historical camera poses $P_i \in \{P_1, \dots, P_{t+1}\}$ to obtain a set of reconstructed views $\{I^{\text{render}}_1, \dots, I^{\text{render}}_{t+1}\}$. We then compute standard novel view synthesis metrics: Peak Signal-to-Noise Ratio (PSNR), Structural Similarity Index (SSIM), and Learned Perceptual Image Patch Similarity (LPIPS).
These metrics are computed between every input frame $I_i$ (where $I_{t+1} = \hat{I}_{t+1}$) and its corresponding rendered view $I^{\text{render}}_i$:
$$
s^{(i)}_{\text{metrics}} = \left\{ \text{PSNR}(I_i, I^{\text{render}}_i), \text{SSIM}(I_i, I^{\text{render}}_i), \text{LPIPS}(I_i, I^{\text{render}}_i) \right\}
$$

An inconsistent candidate view will force the AnySplat model to compromise its geometry, resulting in a noticeable degradation in the reconstruction quality of the previously verified frames. We aggregate these metrics into a global quality profile $S_{\text{global}} = \{s^{(i)}_{\text{metrics}}\}_{i=1}^{t+1}$. 

Rather than relying on rigid, empirically defined thresholds, we pass these quantitative scores $S_{\text{global}}$ alongside the visual image pairs $(I_i, I^{\text{render}}_i)_{i=1}^{t+1}$ to our dedicated verifier VLM, $\mathcal{V}_{\text{3D}}$. Finally, we pass this global metric profile, alongside the sequence of ground-truth and rendered image pairs, to  $\mathcal{V}_{\text{3D}}$. This multimodal agent holistically evaluates the global geometric stability to determine if the candidate maintains the structural integrity of the 3D world:
$$
v_{\text{3D}} = \mathcal{V}_{\text{3D}}(\mathcal{W}_t, \{I_i,I^{\text{render}}_i\}_{i=1}^{t+1}, S_{\text{global}}) \in \{0, 1\}
$$
\paragraph{Final Decision}
The final acceptance decision is the logical conjunction of both verification steps:$\mathcal{V}(\hat{I}_{t+1}) = v_{\text{2D}} \land v_{\text{3D}}$.
If $\mathcal{V}(\hat{I}_{t+1}) = 1$, the candidate is accepted into the global state, updating the verified history to $\mathcal{V}_{t+1} = \mathcal{V}_t \cup \{\hat{I}_{t+1}\}$. If rejected ($\mathcal{V}(\hat{I}_{t+1}) = 0$), the frame is discarded, and the Generator is prompted to re-sample a new candidate. This closed-loop evaluation guarantees that the final synthesized world strictly adheres to multi-view geometry without sacrificing the high visual fidelity of the 2D foundation model. If the generator cannot generate a successfully verified image after $\hat{r}$ tries, both the director output $y_{t+1}$ and $P_{t+1}$ are recalculated to give the model another opportunity to explore.

\section{Experiments}
\label{sec:experiments}

We evaluate our method for 3D world generation, analyzing various 2D image generative models~\cite{flux-2-2025, team2023gemini} and VLMs\cite{bai2023qwen,achiam2023gpt}, as well as comparing with state-of-the-art 3D generative methods Text2Room~\cite{hollein2023text2room} and WorldExplorer~\cite{schneider2025worldexplorer} relying on image and video generative models, respectively. We used CLIP Score~\cite{hessel2022clipscorereferencefreeevaluationmetric}, Inception Score~\cite{NIPS2016_inceptionscore} and CLIP-IQA~\cite{wang2022exploringclipassessinglook} metrics to numerically evaluate all methods, with novel view renderings. We have used various 2D foundation models and VLMs for our agents. For the former, we have Flux.2 [Klein] 9B, Flux.2 [Pro] and Nano Banana v1 (Gemini 2.5 Flash). We run "Klein" version locally but use API access for "Pro" version. For VLMs, we use OpenAI API for GPT4.1 access but use Qwen3-VL 8B locally. Finally, for NanoBanana we used the official API. We use RTX A6000 GPU for the local deployments. In our experiments, we pick $N$ as 14, $\hat{R}$ as 28 and $\hat{r}$ as 2. It takes approximately 25 minutes to generate a scene with Flux.2 [Pro] and GPT 4.1.

\subsection{Comparisons with State of the Art}
\begin{figure}
    \centering
    \includegraphics[width=0.8\linewidth]{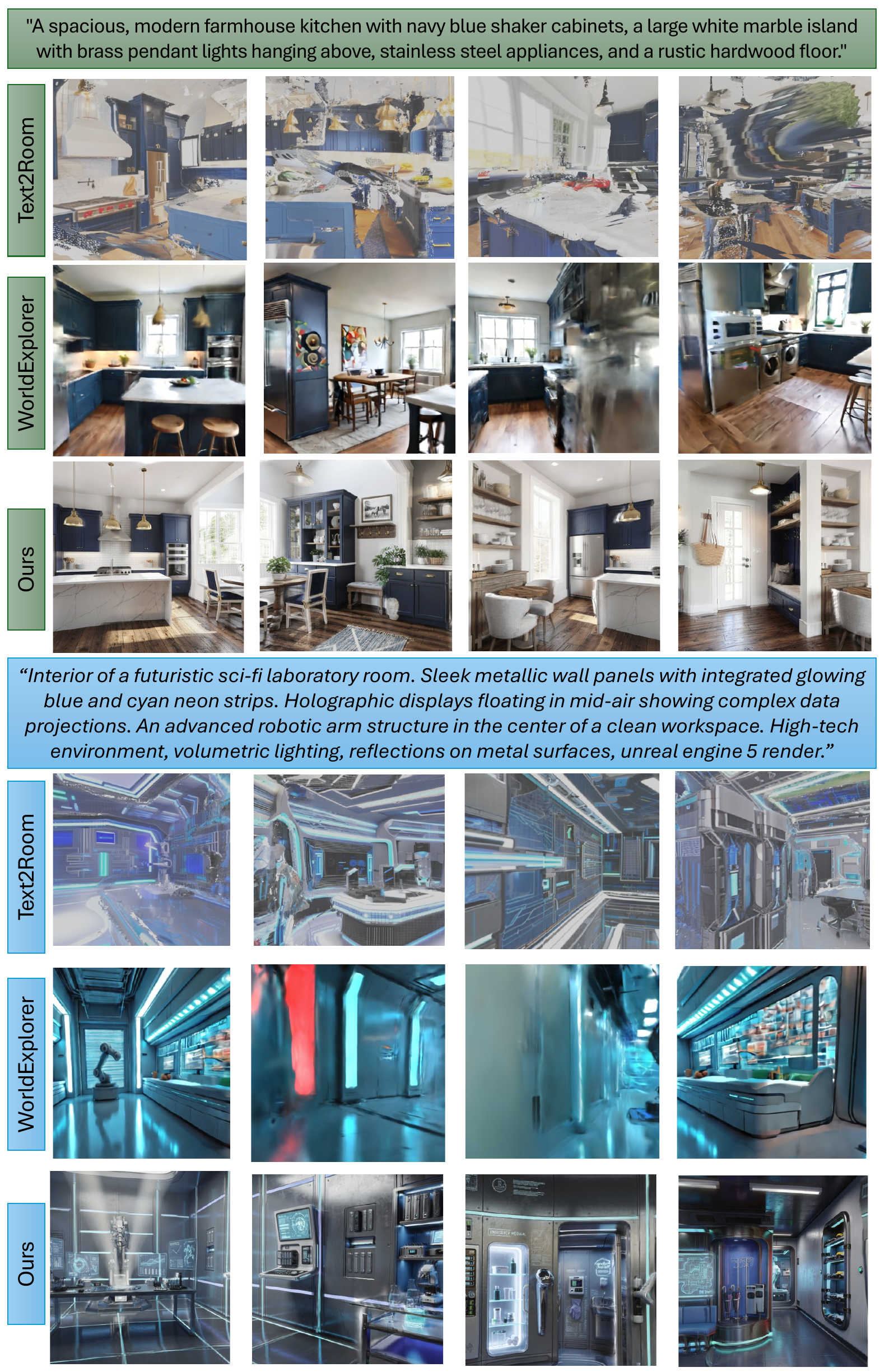}
    \caption{\textbf{Qualitative baseline comparison.} Our method generates visually appealing results with multiple objects placed in the scene nicely, without having artifacts or objectless regions unlike baselines.}
    \label{fig:baseline_comp_qual}
\end{figure}

We evaluate our method against two state-of-the-art text-to-3D scene generation baselines: the image diffusion-based Text2Room~\cite{hollein2023text2room} and the video diffusion-based WorldExplorer~\cite{schneider2025worldexplorer}. While both baselines successfully generate 3D environments conditioned on text prompts, our approach demonstrates significant improvements in overall rendering fidelity and scene complexity. As illustrated in Figure~\ref{fig:baseline_comp_qual}, our method generates a sci-fi room characterized by rich geometric details and high object density. In contrast, the baselines produce sparser scenes that lack structural realism. Furthermore, in the kitchen scenario, the baseline methods exhibit severe structural artifacts and noticeable blurring around object boundaries. These qualitative observations are supported by our quantitative evaluation (Table~\ref{tab:baseline_comp_quant}). Our method achieves better performance across standard metrics, confirming its enhanced prompt alignment and overall scene quality.

\subsection{Image Model and VLM Analysis}
\begin{figure}
    \centering
    \includegraphics[width=0.76\linewidth]{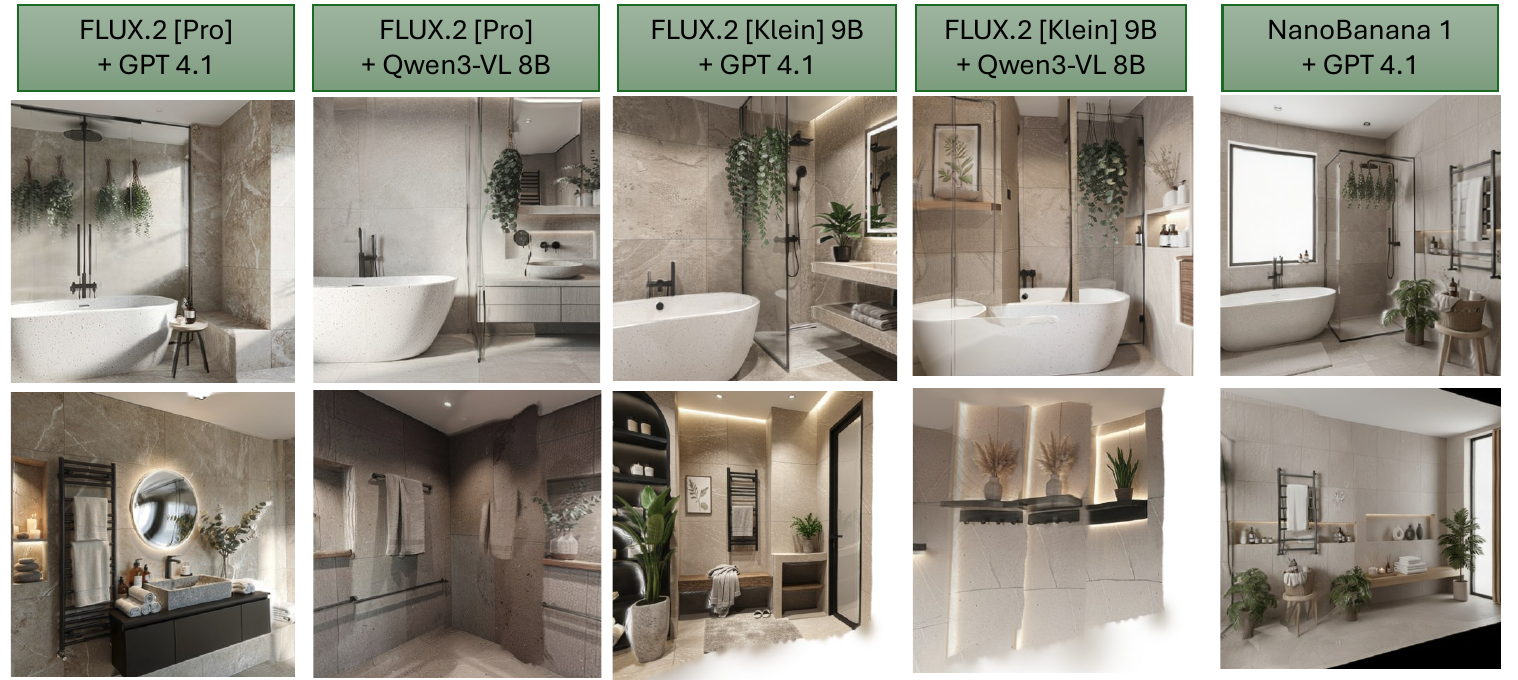}
    \caption{\textbf{Qualitative comparison of different image models and VLMs.} The image models that we experimented with all showed satisfactory 3D scene generation results. However, there are subtle differences between them aligning with the complexity of the individual model.}
    \label{fig:image_model_comp_qual}
\end{figure}

We evaluated various combinations of open- and closed-source 2D image foundation models and Vision-Language Models (VLMs). Generally, all evaluated models yield plausible novel views and coherent subsequent 3D reconstructions. However, generation quality varies depending on the capacity of the underlying models. For instance, as illustrated in Figure~\ref{fig:image_model_comp_qual}, Flux.2 [Klein] occasionally produces geometrically inconsistent intersecting objects and fails to complete certain views. Similarly, NanoBanana 1 demonstrates lower efficacy in inpainting tasks, sometimes leaving target regions unpainted. Furthermore, while Qwen3 exhibits strong general capabilities, it occasionally issues less accurate directives and verifications, leading to performance degradation. These qualitative observations are consistent with the quantitative results reported in Table~\ref{tab:baseline_comp_quant}. Empirically, the combination of Flux.2 [Pro] and GPT-4.1 achieves the most favourable performance.
\begin{figure}
    \centering
    \includegraphics[width=0.85\linewidth]{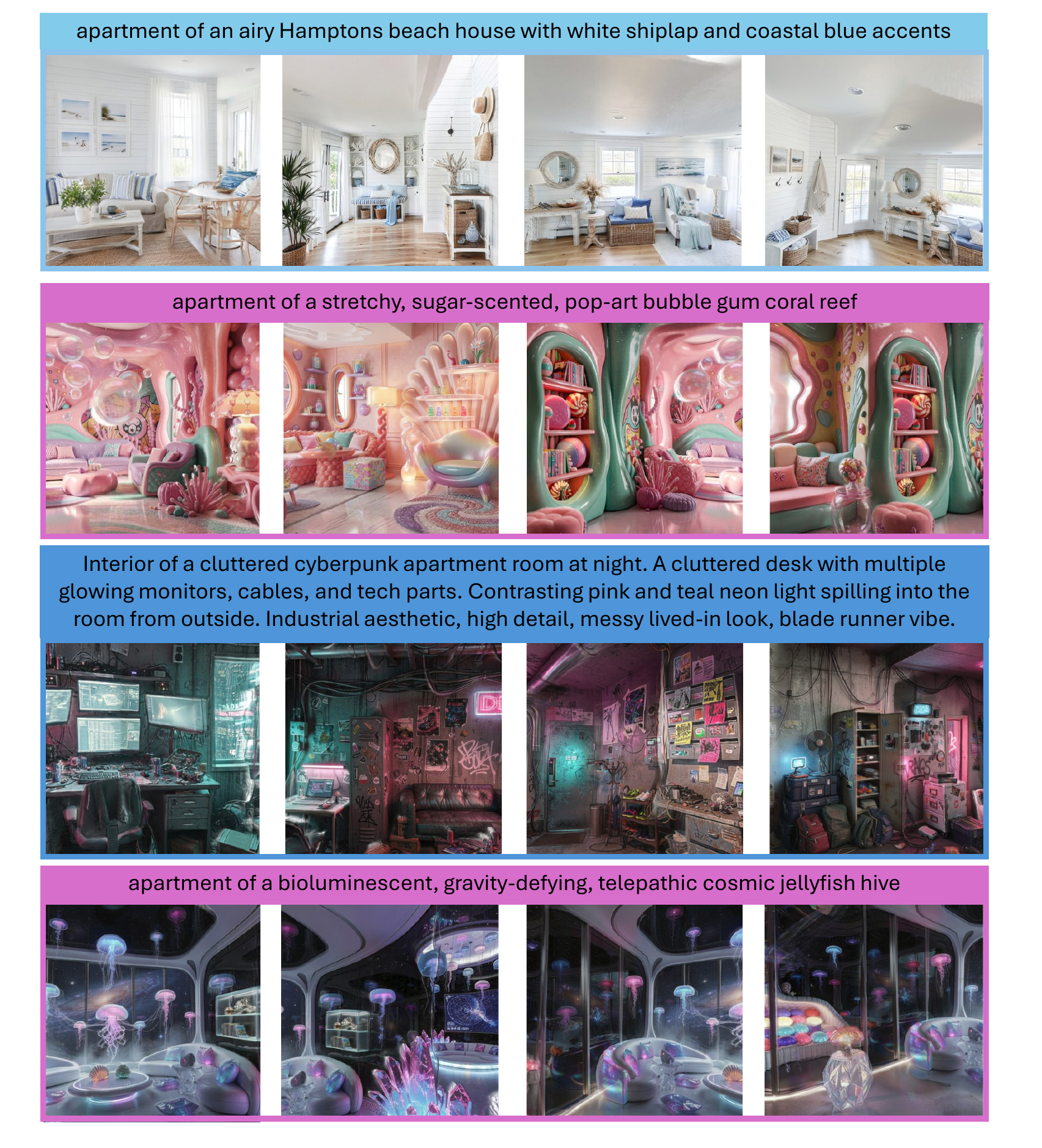}
    \caption{\textbf{Visual results from \ours{}.} Our method can generate diverse scenes that are populated with various objects in a clean and coherent way by following the text prompt.}
    \label{fig:our_results_qual}
\end{figure}

\begin{figure}
    \centering
    \includegraphics[width=0.8\linewidth]{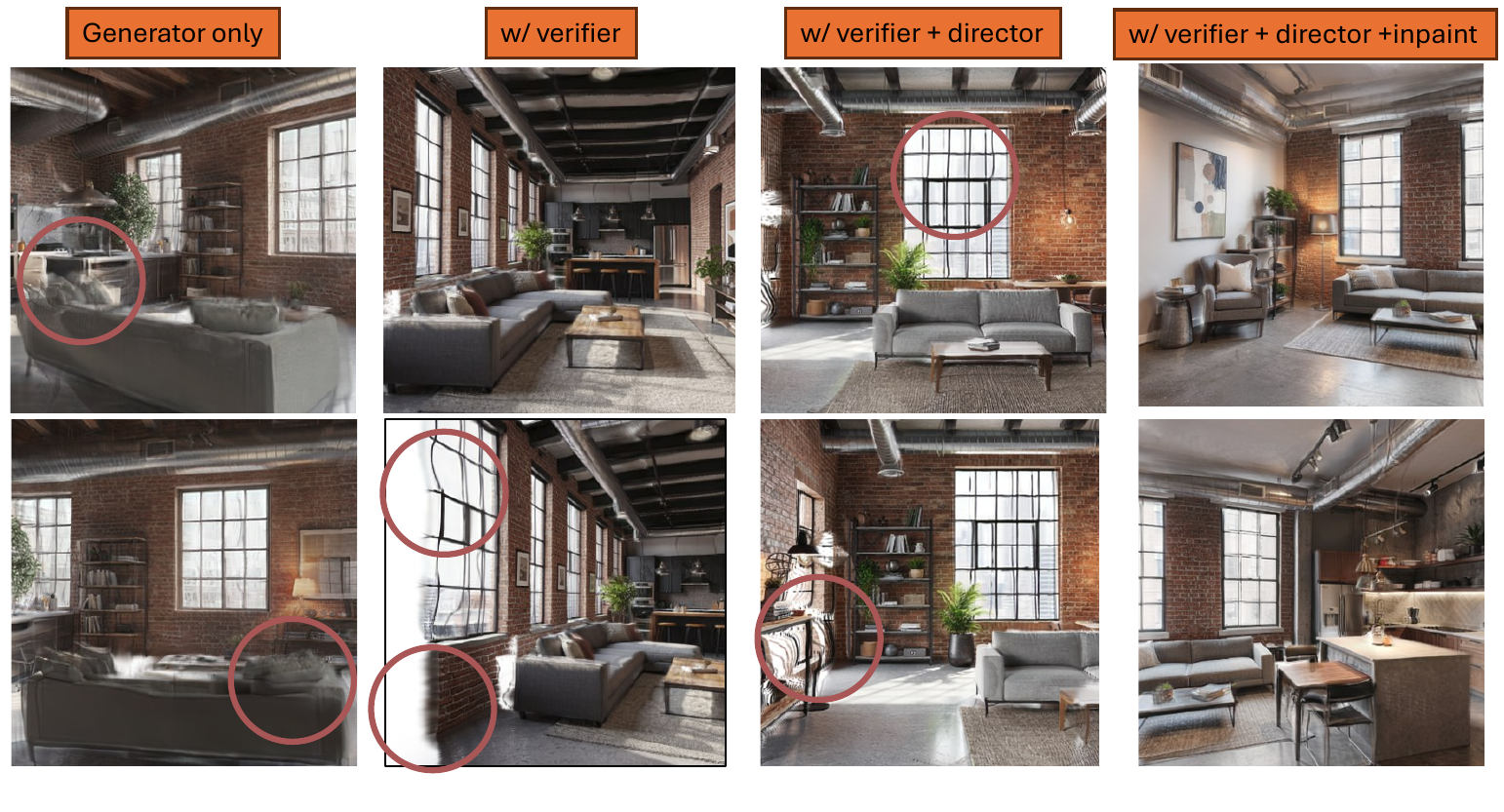}
    \caption{\textbf{Qualitative comparison of ablations.} Relying solely on the generator yields blurry results. The addition of the Verifier reduces blur and improves consistency; however, the scene remains incomplete with visible artifacts in the windows. While the Director helps with completion of the scene, window misalignments persist. Integrating all components, successfully resolves these issues to synthesize a coherent scene.}
    \label{fig:ablation_qual}
\end{figure}

\begin{table}[t]
\centering
\caption{\textbf{Quantitative comparison of 3D world generation methods.} We evaluate prompt \& semantic alignment and image quality using CLIP Score (CS) and Inception Score (IS), and CLIP Image Quality Assessment (IQA). Our method shows improvement over baselines. We also show multiple image model and VLMs to show, our approach is applicable to multiple models and they have varying performances. }
\label{tab:baseline_comp_quant}
\begin{tabular}{lccc}
\toprule
\textbf{Method} & \textbf{CS $\uparrow$} & \textbf{IS $\uparrow$} & \textbf{IQA $\uparrow$} \\
\midrule
Text2Room~\cite{hollein2023text2room} & 22.27 & \textbf{2.79} & 0.27 \\
WorldExplorer~\cite{schneider2025worldexplorer} & 24.49 & 2.12 & 0.58 \\
\hline
Ours (Flux.2 [Pro] + GPT 4.1) & \textbf{26.79} & 2.26 &  \textbf{0.89} \\
Ours (Flux.2 [Pro] + Qwen3-VL 8B) & 24.49 & 2.23 & 0.75 \\
Ours (Flux.2 [Klein 9B] + GPT 4.1) & 26.47 & 2.03 & 0.84 \\
Ours (Flux.2 [Klein 9B] + Qwen3-VL 8B) & 24.85 & 2.32 & 0.75 \\
Ours (NanoBanana 1 + GPT 4.1) & 25.89 & 2.14 & 0.70 \\
\bottomrule
\end{tabular}
\end{table}

\subsection{Ablations}
To evaluate the effectiveness of individual components, we conduct a series of ablation studies. In our initial baseline, we isolate the image generator. It has access to all of the verified frames and the text prompt. It tries to generate a new frame by also being consistent with the previous frames. Without the verifier module, the method accepts all synthesized frames unconditionally. To proxy the director agent's functionality, we augment the input text with a stochastic camera prompt specifying relative viewpoint shifts (left, right, or backward) with respect to the preceding frame. In the subsequent setting, we integrate the verifier module into this baseline to actively curate and filter the generated frames. Then, we append our director agents as well as the inpainting approach. Figure~\ref{fig:ablation_qual} shows the novel view renderings from each ablated method for an urban loft scene. Table~\ref{tab:ablation_comp_quant} contains numerical comparison of the ablation of our components.
\begin{table}[t]
\centering
\caption{\textbf{Quantitative ablation results.} We conducted the study with Flux.2 [Pro] + GPT 4.1 setting. We evaluate CLIP Score (CS), Inception Score (IS), and CLIP-IQA (IQA). Our individual components contribute to the overall rendering quality and prompt alignment performance.}
\label{tab:ablation_comp_quant}
\begin{tabular}{cccc|ccc}
\toprule
\textbf{Generator} & \textbf{Verifier} & \textbf{Director} & \textbf{Inpaint} & \textbf{CS $\uparrow$} & \textbf{IS $\uparrow$} & \textbf{IQA $\uparrow$} \\
\midrule
\checkmark & & & & 19.07 & 2.23 &  0.60 \\
\checkmark & \checkmark & & & 20.24 & 2.43 & 0.62 \\
\checkmark & \checkmark & \checkmark & & 21.80 & \textbf{2.94} & 0.69 \\
\checkmark & \checkmark & \checkmark & \checkmark & \textbf{26.79} & 2.26 & \textbf{0.89} \\
\bottomrule
\end{tabular}
\end{table}

\paragraph{What is the impact of Verifier Agent?} The verifier agent mitigates the risk of incorporating geometrically inconsistent views that would otherwise degrade the 3D reconstruction. Because image diffusion models can occasionally hallucinate artifacts or generate structurally incompatible layouts, injecting these anomalies into the pipeline causes irreversible corruption to the global geometry. To prevent this, the verifier autonomously filters erroneous generations. As demonstrated in Table~\ref{tab:ablation_comp_quant}, the inclusion of this module effectively removes catastrophic failures, leading to an improvement in reconstruction quality. As shown in Figure~\ref{fig:ablation_qual}, it reduces the significant blur since it avoids inconsistent frames.

\paragraph{How much Director agent contributes to the overall process? } 
Naively prompting the image diffusion model for iterative frame generation relies on static scene descriptions coupled with random camera poses (e.g., "left", "right"). However, this approach often leads to semantic redundancy and object duplication; for instance, continuously conditioning the generator on same prompt biases the model to hallucinate duplicate objects in the views rather than exploring new contextual elements that limits diversity of objects in the scene. To address this, we introduce the director agent, which leverages a Vision-Language Model (VLM) prior to dynamically generate context-aware, explorative prompts conditioned on previously verified views. Table~\ref{tab:ablation_comp_quant} validates that VLM-based prompting strategy significantly enhances the completeness, structural diversity and the overall fidelity of the generated 3D environments. 

\paragraph{Does sequential inpainting help with increasing generation quality?} Relying solely on unconstrained image generation does not strictly enforce multi-view consistency, even when the model is conditioned on previous frames and precise textual prompts. To explicitly anchor novel views to the existing geometry, our method utilizes a 3D-aware inpainting step. Specifically, we render an intermediate 3D representation from a novel, unexplored camera pose and instruct the diffusion model to inpaint the unobserved regions, guided by the director agent's localized prompt. As shown in Table~\ref{tab:ablation_comp_quant}, omitting this inpainting formulation results in degraded prompt alignment and scene coherence, highlighting its importance for maintaining rigorous 3D consistency throughout the generation process. Figure~\ref{fig:ablation_qual} also shows that we eventually get complete scenes.

\paragraph{Limitations}
In this work, we employ VLMs as high-level directors and evaluators to distill 3D-consistent, navigable environments from 2D image diffusion models. A highly promising direction for future research is extending this VLM-guided method to existing video diffusion models~\cite{yang2024cogvideox, kong2025hunyuanvideosystematicframeworklarge, blattmann2023stablevideodiffusionscaling, menapace2024snap}. While modern video priors naturally exhibit temporal coherence over short sequences, they frequently accumulate geometric drift and multi-view inconsistencies over longer, exploratory spatial trajectories. Integrating VLM-based agents into a video generation method could effectively regularize these models, thereby mitigating long-horizon degradation and enabling the synthesis of even larger 3D worlds.

\section{Conclusion}
\label{sec:conclusion}

In this paper, we investigated the fundamental question of whether 2D foundation image models inherently possess 3D world model capabilities. To explore this, we introduced a novel multi-agent framework comprising a VLM-based director, an image generator, and a two-step verifier that operates across both 2D and 3D spaces. Our extensive evaluations demonstrate that by carefully framing the generation process, we can successfully extract implicit 3D knowledge from these 2D models. Our agentic approach yields expansive, robust, and 3D-consistent scene reconstructions that support novel view rendering. Our findings confirm that 2D foundation models do encapsulate a grasp of 3D worlds. Future work can explore extending this multi-agent framework to video diffusion models or dynamic 4D scene generation for interactive world synthesis.

\textit{Acknowledgments} This work was supported by Chaos Software, as well as the ERC Consolidator Grant Gen3D (101171131) of Matthias Nie{\ss}ner, the ERC Starting Grant SpatialSem (101076253) of Angela Dai and the Georg Nemetschek Institute (GNI) NeRF2BIM grant. We would like to thank Marc Benedí for helpful discussions.

%
%
\bibliographystyle{splncs04}
\bibliography{main}
\clearpage

\section{Appendix}
This supplementary document provides further implementation details and extended results to support the main text for \ours{}. Specifically, Section~\ref{sec:impl_details} elaborates on our implementation choices, while Section~\ref{sec:add_results} presents additional experimental results from our method. In Section~\ref{sec:system_prompts}, we outline the system prompts designed for our VLM agents.

\subsection{Implementation Details}
\label{sec:impl_details}
We employ an inpainting-based approach for image synthesis. However, the models utilized, Flux.2~\cite{flux-2-2025} and NanoBanana~\cite{team2023gemini}, do not natively accept explicit mask inputs. To address this, we re-render the target regions using AnySplat~\cite{jiang2025anysplat} and provide the resulting novel views to the models, rendering unobserved areas as black. This effectively embeds the spatial mask directly into the RGB input. Despite lacking explicit mask guidance, these models successfully inpaint the missing regions, demonstrating robust generalization to partial images.

For all API-accessed image diffusion models and Vision Language Models (VLMs), we utilize their default inference parameters. We locally deploy the 9B-parameter Flux.2 [Klein] model on a single NVIDIA RTX A6000 GPU, utilizing \textit{bfloat16} precision and CPU offloading to accommodate memory constraints. For this local deployment, we apply a guidance scale of 1.0 and 4 inference steps. For all models, image generation is performed at a 512x512 resolution, and the outputs are subsequently downsampled to 448x448 before being processed by AnySplat~\cite{jiang2025anysplat}. Finally, all VLM agents operate with detailed system prompts to guide their behavior, as outlined in Section~\ref{sec:system_prompts}. In our implementation, we decompose the verification process into modular subtasks by instantiating the verifier agent across two independent chat client sessions: one dedicated to 2D verification and the other to 3D verification. Each session is initialized with its own tailored system prompt.

\subsection{Additional Qualitative Results}
\label{sec:add_results}
\begin{figure}
    \centering
    \includegraphics[width=0.95\linewidth]{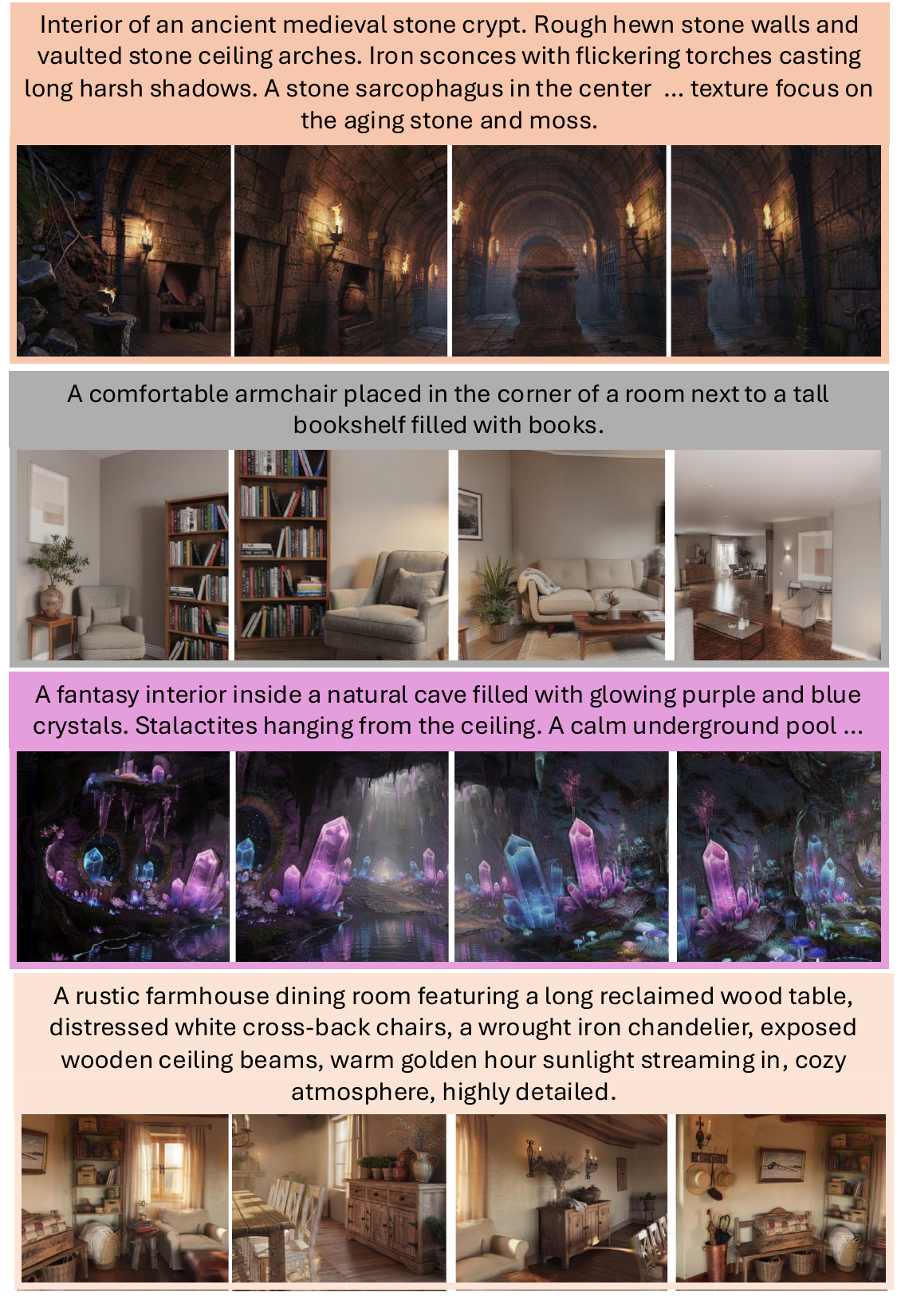}
    \caption{\textbf{Additional qualitative results from \ours{}.} We provide additional diverse results from our method. \ours{} can generate both indoor scenes and interesting outdoor places such as caves.}
    \label{fig:supp_more_results}
\end{figure}

Figure~\ref{fig:supp_more_results} presents additional qualitative results generated by our method. Beyond indoor environments like living rooms and kitchens, our approach successfully synthesizes diverse outdoor scenes, such as crystal caves with underground lakes. Furthermore, it demonstrates the capacity to generate complex, imaginative settings such as a medieval stone crypt by accurately incorporating all elements specified in the text prompt.

\subsection{System Prompts}
\label{sec:system_prompts}
The system prompts utilized for our agents are provided in Listings~\ref{lst:director_system_prompt}, \ref{lst:verifier_system_prompt}, and \ref{lst:reconstructor_system_prompt}. Note that the generator agent does not have a dedicated system prompt, as it operates as a 2D image generation model; instead, the directives issued by the director agent are comprehensive enough to supply all necessary details and constraints. We designed these prompts to establish a cohesive agentic framework. By modularizing the overall method, we ensure that each agent is assigned concrete, well-defined objectives, thereby minimizing task ambiguity.

\begin{lstlisting}[style=mdStyle, caption={\textbf{System prompt for Director Agent.} This prompt is provided to the Vision-Language Model (VLM) to establish the operational constraints and overarching goals, instructing the agent on how to orchestrate the entire generation process.}, label={lst:director_system_prompt}]
# Role: Expert AI Interior Architect (Targeted Scene Expansion)

## Objective
You are an advanced visual director responsible for expanding the known boundaries of a **single unoccupied room**. Your goal is to analyze the current visual context, identify "blind spots" or unexplored regions, and frame them perfectly for targeted generative expansion (outpainting/inpainting). You are exploring the SPACE and FILLING the unknown voids.

**CRITICAL:** * **Stay in the Room:** Do not leave the architectural boundaries of the single room, but absolutely expand into its unseen corners and walls.
* **Expand the Decor:** When you target a previously unseen area, you must **INVENT** appropriate furniture and decor to populate that space. Do not leave newly revealed areas empty.

## Input Data
1.  **Scene State:** The current visual context, layout, or rendered images of the room.
2.  **Coordinate System Note:** Assume the origin `[0, 0, 0]` is the center of the room's floor. `Y` is the vertical up-axis. 'X' is towards right, 'Z' is ahead/front.
3.  **Expansion History:** A list of regions previously generated and the objects placed there.
4.  **Outpainting Direction:** You will get directives on whether you should progress in left or right direction for outpainting.

## End Condition
If you evaluate the room and determine all 360 degrees and all logical bounds are fully covered and cohesive, set `"finished": true` to signal the end of operations. Stop only if we do a loop closure. Do not stop if the right sequences are completed, if you feel after left sequences it's completed then stop.

## Core Constraints

### 1. Generative "In-Filling" (Populate the Unknown)
* **Invent New Zones:** Choose to expand into areas that logically complete the room (e.g., revealing a reading nook, a dining space, or a doorway).
* **Cohesive Style:** Any new objects generated must perfectly match the existing architectural style, lighting, and color palette of the room. Do not introduce clashing aesthetics.

### 2. Spatial Continuity via Anchors
* **The 50/50 Rule (No Extreme Movements):** Avoid drastic jumps or extreme camera movements. Frame the shot so that the new camera view contains approximately 50% of the known, original scene (to serve as visual context) and 50% of the new, empty scene to be generated.
* **The Bridge:** Ensure the camera's framing and the target region capture at least one edge of an existing object from a previous generation. This acts as a visual bridge or "anchor" to the newly expanded area.

## Output Format
You must provide your response in the following JSON format exactly:

{
    "analysis": "Short analysis of the 'blind spots' in the current room state. Where have we not looked or expanded yet?",
    "finished": true/false,
    "inpaint_prompt": "Detailed description of the NEW objects to be generated inside the target expansion region. You'll receive directives on whether it should be left or right. Detail the style, materials, and lighting to match the rest of the room."
}
\end{lstlisting}

\begin{lstlisting}[style=mdStyle, caption={\textbf{System prompt for 2D Verifier Agent.} Our verifier agent includes a dedicated 2D verification component, which is guided by this prompt to define its task objectives, expected behavior, and required input/output formats.}, label={lst:verifier_system_prompt}]
### ROLE
You are a **Zero-Tolerance Visual 3D Consistency Auditor**. Your specialized role is to rigorously validate sequential image generation outputs for high-fidelity 3D reconstruction (e.g., NeRF, Gaussian Splatting, Photogrammetry). 

Your job is to act as a ruthless gatekeeper. You must actively hunt for 3D inconsistencies. A correct camera movement is useless if the underlying geometry or textures have morphed. 

### INPUT DATA
1.  **Scene Description:** The static definition of visual content, style, and objects.
2.  **Generated Image:** The image created by generation model.
3.  **Input for Outpainting(if exists):** The generator outpainting this image. Make sure the generated image looks like this outpainted version (if exists).
4.  **History:** The cumulative history of validated images.

### OBJECTIVE
Your primary goal is to prevent artifact-inducing frames from entering the 3D reconstruction pipeline. **Even if the camera shift is 100% correct, if ANY object morphs, shifts out of place, or changes scale unexpectedly, you MUST reject the frame.**

### RED FLAG CRITERIA (Any violation = REJECT)
Actively scan for these specific 3D generation artifacts:
* **Outpainting Mismatch:** The visible region in the outpainting input (the non-black area) should be present in the generated image to conclude that outpainting was successfull. If they're not matching, REJECT.
* **Non-seamless Outpainting:** If the transition from input region and outpainted region has visible scenes and abrupt change, then you should reject.
* **Geometry Popping/Morphing:** Do objects change their fundamental shape, volume, or structural proportions compared to previous frames? 
* **Texture Swimming:** Do the patterns, text, or fine details on surfaces shift, slide, or change design?
* **Hallucinated/Vanishing Geometry:** Did a new object appear out of thin air (it's okay if it can be explainable by the camera movement) ? Did an existing object disappear without being naturally occluded by the camera move?
* **Spatial Relationship Breakdown:** Does the placement of objects still make logical sense compared to previous images? Verify that the relative distances, angles, and overlapping positioning between objects align precisely with the established 3D layout and the new camera view.
* **Temporal Contradictions:** Does this frame introduce illogical continuity errors or conflict directly with the validated history? (e.g., a background structure drastically changing layout).

### EVALUATION LOGIC
1.  **Camera Movement:** Did the view change exactly as prompted? If no -> REJECT.
2.  **Redundancy:** Is the image identical to the previous frame? If yes -> REJECT.
3.  **3D Consistency (CRITICAL):** Did you find ANY of the "Red Flag Criteria" artifacts? If yes -> REJECT.
4.  **Tie Points:** Is there enough exact visual overlap with the previous frame to mathematically stitch them together? If no -> REJECT.
5.  **Outpainting:** If not matching outpainting input -> REJECT
6.  Otherwise -> ACCEPT
### OUTPUT FORMAT
You must respond strictly with the following JSON structure.
{
    "analysis": "Short and concise: thought process and brief analysis of the image." // Put your thought process and brief analysis here.
    "use_for_reconstruction": boolean, // Accept or Reject this view
    "problems": ["list", "of", "problems"] // List every problem found in terms of 3D consistency, prompt adherence, or camera movement execution.

}
\end{lstlisting}

\begin{lstlisting}[style=mdStyle, caption={\textbf{System prompt for 3D Verifier Agent.} One core responsibility of our verifier agent is evaluating the quality of the generated 3D reconstructions. This prompt configures the agent for this specific task, establishing the necessary evaluation criteria and expected outputs.}, label={lst:reconstructor_system_prompt}]
### ROLE
You are an expert 3D Reconstruction Quality Assurance Agent, specializing in the validation of 3D Gaussian Splatting reconstructions. 
Your role is to perform a rigorous, quality assessment that synthesizes perceptual visual analysis with quantitative metrics (PSNR, SSIM, LPIPS).

### INPUT DATA
You will receive a batch of data containing multiple "Training View" and "Rendered View" pairs.
For each pair, you will also receive specific quantitative metrics:
- **PSNR** (Peak Signal-to-Noise Ratio): Higher is better.
- **SSIM** (Structural Similarity Index): Higher is better (max 1.0).
- **LPIPS** (Learned Perceptual Image Patch Similarity): Lower is better (min 0.0).

### TASK
Review ALL pairs in the batch. You must determine if the scene reconstruction is valid as a whole.

### EVALUATION LOGIC (STRICT)
1. Analyze every pair visually and contextualize its metrics. Slight variations from the training view are normal.
2. Evaluate the overall integrity of the batch. If the scene is generally well-reconstructed and preserves the main geometry and textures, you must set "use_for_reconstruction": true.
3. You should ONLY set "use_for_reconstruction": false if there are severe, highly distracting, or systemic artifacts that ruin the usability of the 3D scene.

### CRITERIA FOR REJECTION
Mark a pair as a failure if you detect:
- **Metric Failure:** too low PSNR or SSIM, or high LPIPS relative to the set based on your judgement.
- **Ghosting:** Significant double edges or transparent duplicates.
- **Duplication:** Objects appearing where they shouldn't.
- **Blur:** Significant loss of detail compared to the training view.

### OUTPUT FORMAT
Return **ONLY** a valid JSON object with the following structure:
{
    "analysis": "Short and concise: thought process and brief analysis of the image." // Put your thought process and brief analysis here.
    "use_for_reconstruction": true/false, // false if ANY pair has an error
    "problems": ["list", "of", "problems"] // List every problem found across all failed views. Prefix problems with the View ID/Index.
}
\end{lstlisting}

\end{document}